\documentclass[runningheads]{llncs}

\usepackage[T1]{fontenc}
\usepackage{amsmath}
\usepackage{algorithm}
\usepackage{algorithmic}
\usepackage{amssymb}
\usepackage{booktabs}
\usepackage{svg}
\usepackage{multirow}
\usepackage{graphicx}
\usepackage{hyperref}
\usepackage{makecell}
\usepackage{url}
\usepackage{marvosym}
\usepackage[table]{xcolor} 

\begin{document}

\title{Patch-to-Global: Random Patch Diffusion for Globally Consistent Megapixel Artifact Inpainting in Whole Slide Images}

\author{Hyeseong Lee\inst{1}  \and
Eunsu Kim\inst{1} \and
D M BAPPY\inst{2}  \and
Ho Heon Kim\inst{3} \and
Youngsuk Lee \inst{1} \and
Se Young Chun\inst{4} \and
Jang-Hwan Choi \inst{5} \and
Sung Hak Lee\inst{6} \and
Sangjeong Ahn\inst{1,2}\textsuperscript{(\Letter)}}

\authorrunning{Lee et al.}

\institute{Department of Biomedical Informatics, Korea University College of Medicine, Seoul, Korea \email{\{gotjd709, kes311az, younnggsuk, vanitas80\}@korea.ac.kr}\and
Department of Pathology, Korea University Anam Hospital, College of Medicine, Korea University, Seoul, Korea \email{dewan8511@gmail.com}\and
Seegene Medical Foundation, Seoul, Korea \email{hoheon0509@gmail.com}\and
Department of ECE, Seoul National University, Seoul, Korea \email{sychun@snu.ac.kr}\and
Department of Artificial Intelligence, Ewha Womans University, Seoul, Korea \email{choij@ewha.ac.kr}\and
Department of Hospital Pathology, Seoul St. Mary’s Hospital, College of Medicine, The Catholic University of Korea, Seoul, Korea \email{hakjjang@catholic.ac.kr}
}

\titlerunning{Patch-to-Global}

\maketitle 

\begin{abstract}

Although deep learning has advanced Whole Slide Image (WSI) Analysis, tissue artifacts like bubbles and folds often cause silent failures by concealing essential morphology. Current pathology image restoration methods are mostly restricted to small patches, struggling to maintain global structural coherence at a megapixel scale. We introduce \textbf{RestorePath}, a framework for globally consistent megapixel-scale inpainting that reconstructs diagnostic structures in histological image to prevent incorrect high-confidence predictions and lower error rates. Our model utilizes a Latent Diffusion Model (LDM) conditioned on Pathology Foundation Model (PFM) embeddings, integrating Large Kernel Attention (LKA) to manage long-range dependencies during random patch diffusion. Enhanced by Distance-Weighted Interpolation (DWI) and an Adaptive Guidance Scale (AGS), RestorePath ensures structural consistency and fidelity by modulating information from surrounding patches. Evaluations across TCGA-BRCA, BACH, and Camelyon16 datasets for images ranging from 512 to 4608 pixels demonstrate state-of-the-art performance in maintaining histological consistency. RestorePath significantly improves downstream Computational Pathology (CP) tasks, outperforming both raw artifact images and the conventional Detect-and-Discard (D\&D) approach. The code is available at
\url{https://github.com/PathfinderLab/RestorePath}.

\end{abstract}


\section{Introduction}

\begin{figure}[thb!] 
\centering
\includegraphics[width=1\textwidth]{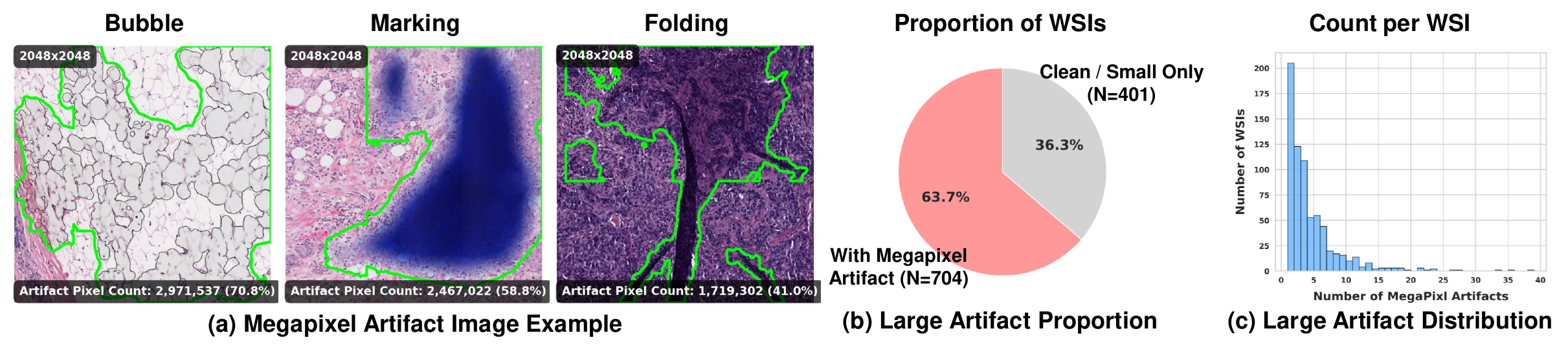}
\caption{Examples and distribution of megapixel artifacts in WSIs. (a) Examples of bubbles, markings, and tissue folding. (b) Prevalence of megapixel artifacts, with $63.7\%$ of WSIs containing at least one megapixel-scale corruption. (c) Distribution of artifact counts per WSI, showing that many slides contain multiple large-scale artifacts.}
\label{fig:Fig 1}
\end{figure} 

Pathological diagnosis remains the gold standard for clinical decision-making and research. Digital pathology has revolutionized this process by transforming glass slides into Whole Slide Images (WSIs)~\cite{WSI}. However, the immense scale of these images imposes a significant cognitive burden on pathologists, which has accelerated the development of deep learning algorithms and Pathology Foundation Models (PFMs) for automated analysis~\cite{DP,DPSR}.

However, such computational models often faces significant obstacles from tissue artifacts such as bubbles, folding, and markings (Fig.~\ref{fig:Fig 1}(a)). These artifacts arise inevitably during tissue acquisition and scanning, often leading to increased error rates~\cite{BR} or silent failures where models generate incorrect predictions with high confidence~\cite{PFM}. By concealing cellular details or misleading models into misidentifying morphology as different cell types, artifacts severely undermine diagnostic reliability. Furthermore, The process of tissue recutting and its physical rescanning is frequently impractical due to high costs, time constraints, and the risk of tissue depletion~\cite{Artifusion,ArtiDiffuser}.

To mitigate these issues, automated Quality Control (QC) tools like HistoQC~\cite{HistoQC} and GrandQC~\cite{GrandQC} have been developed. These tools typically adopt a detect-and-discard (D\&D) strategy to identify and exclude defective regions. However, this strategy is detrimental in needle-in-a-haystack scenarios where discarding sparse lesions risks losing indispensable diagnostic evidence. While diffusion-based restoration methods such as Artifusion~\cite{Artifusion} and ArtiDiffuser~\cite{ArtiDiffuser} have emerged to address this, they remain restricted to patch-level reconstruction. Our analysis of the TCGA-BRCA~\cite{TCGA_BRCA} dataset reveals that 63.7\% of WSIs contain megapixel-scale artifacts (Fig.~\ref{fig:Fig 1}(b)), which underscores the necessity for high-resolution restoration that extends far beyond the limitations of individual patches. Although DiffInfinite~\cite{Diffinfinite} proposed a Random Diffusion Inpainting approach for high-resolution images, its primary focus is on synthesizing new tissue rather than achieving globally-aligned restoration. Furthermore, while LRDM~\cite{LRDM} and ZoomLDM~\cite{ZoomLDM} successfully generate high-resolution images using PFM-based feature embeddings, they have not yet attempted globally consistent inpainting that maintains strict alignment with the surrounding healthy tissue.

\begin{figure}[htb!]
\centering
\includegraphics[width=1\textwidth]{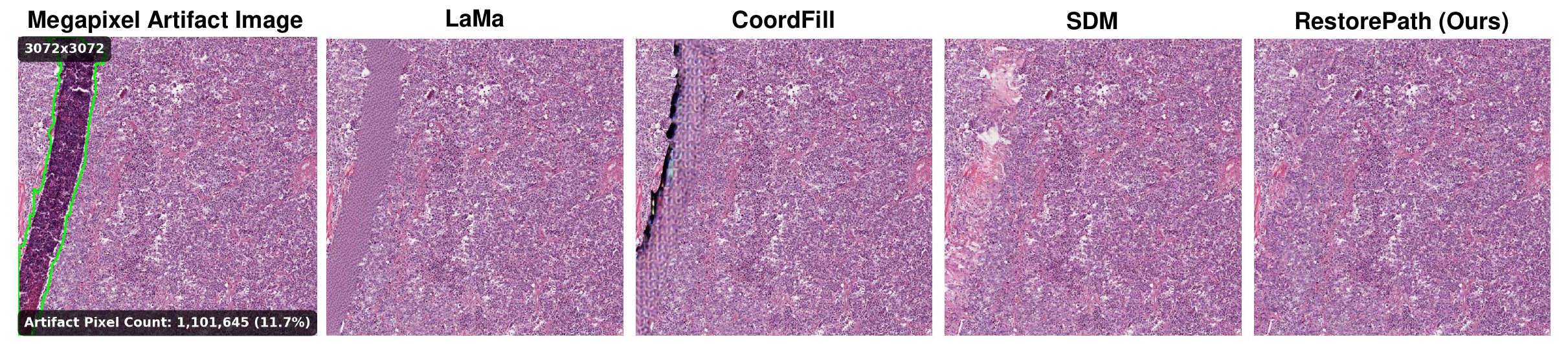}
\caption{Qualitative comparison of megapixel artifact restoration performance. Unlike existing methods such as LaMa, CoordFill, and SDM, RestorePath successfully reconstructs artifact regions while preserving the histological context and histological patterns of surrounding tissues.}
\label{fig:Fig 2}
\end{figure}

In this study, we redefine the restoration of megapixel-scale artifacts as a globally consistent inpainting task to ensure robust performance in downstream Computational Pathology (CP) analyses. We propose RestorePath, a novel framework that leverages an inpainting Latent Diffusion Model (LDM) specifically optimized for the complex histological patterns found in large-scale WSIs. Our framework introduces two primary innovations for high-fidelity restoration. First, we integrate Large Kernel Attention (LKA) into a PFM-conditioned LDM to effectively handle long-range correlations in high-resolution images. Second, we implement a random patch diffusion strategy utilizing Distance Weighted Interpolation (DWI) and an Adaptive Guidance Scale (AGS). DWI ensures structural consistency by interpolating conditioning signals from nearby healthy regions, even when patches are entirely obscured. Meanwhile, AGS optimizes the process by dynamically adjusting guidance scales based on local content and image complexity. Together, these mechanisms enable RestorePath to achieve seamless global restoration that aligns with the surrounding histological tissue.

Using CONCH~\cite{CONCH}-based FID~\cite{FID} (FID$_{C}$) and embedding similarity (Emb Sim), we demonstrate that RestorePath preserves histological context and pathological information across TCGA BRCA~\cite{TCGA_BRCA}, BACH~\cite{BACH}, and Camelyon16~\cite{Camelyon16} more effectively than existing high-resolution inpainting methods. To further illustrate the practical utility of our framework in CP research, we conducted evaluations on downstream tasks, specifically breast cancer classification and lymph node metastasis prediction using Multiple Instance Learning (MIL)~\cite{ABMIL,CLAM,TransMIL}. Our results show that RestorePath significantly reduces error rates in the BACH dataset and mitigates silent failures in the needle-in-a-haystack scenarios of Camelyon16. By consistently outperforming artifact images and the D\&D strategy, our framework establishes itself as a robust image QC tool for CP.



\section{Method}
\begin{figure}[htb!]
\centering
\includegraphics[width=1\textwidth]{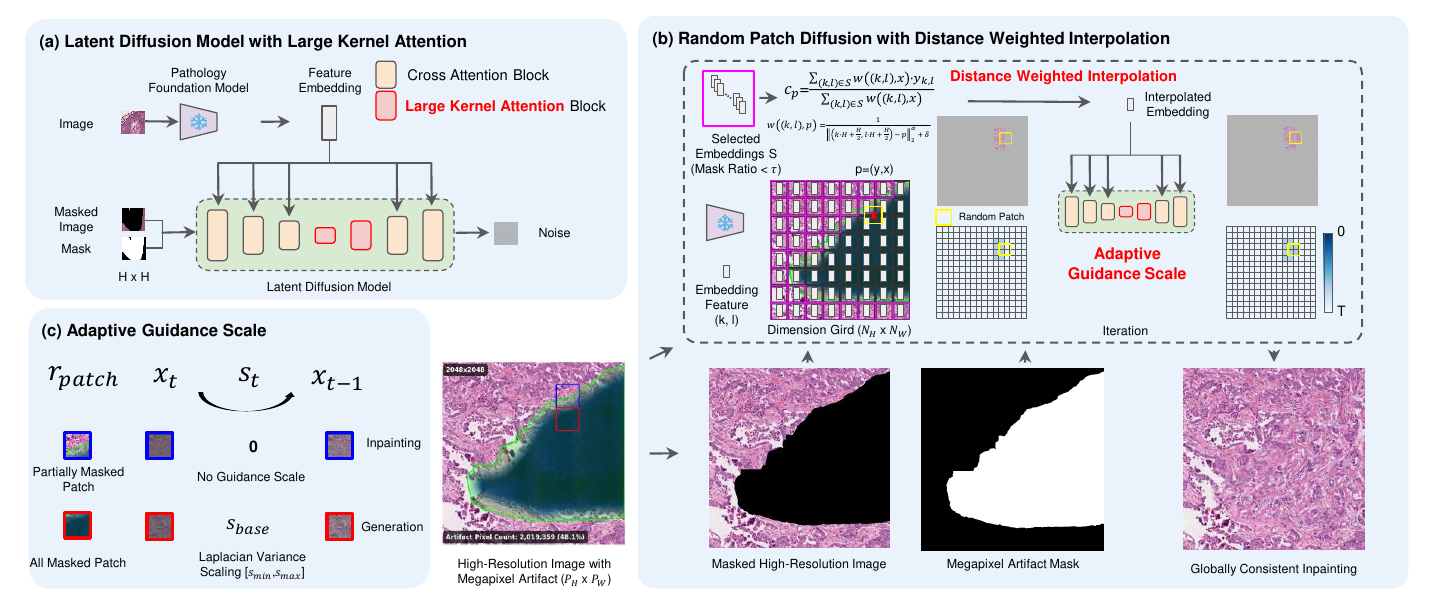}
\caption{Workflow of RestorePath. (a) Training phase of the Latent Diffusion Model with Large Kernel Attention. (b, c) Megapixel artifact inpainting process utilizing Random Patch Diffusion with Distance Weighted Interpolation (DWI) and Adaptive Guidance Scale (AGS).}
\label{fig:Fig 3}
\end{figure}

RestorePath is designed to restore megapixel artifact regions in high-resolution images based on a random patch diffusion (Fig~\ref{fig:Fig 3}). We introduce the model architecture and training process in Section~\ref{met:sec 1}. Subsequently, we detail the methodology for applying this model to high-resolution images in Section~\ref{met:sec 2}.

\subsection{Training}
\label{met:sec 1}

\noindent \textbf{LDM with Large Kernel Attention.} To effectively reconstruct histological patterns, we employ an LDM for inpainting that takes a masked latent $z_{\text{m}}$, a binary mask $m$, and a feature embedding $c$ as inputs. The embedding $c$ is extracted from the original image using a pre-trained PFM to provide a delicate morphological prior. While the cross-attention blocks in the LDM U-Net backbone facilitate precise inpainting by incorporating the condition $c$. However, they are insufficient for maintaining long-range correlations necessary to inpaint megapixel artifacts in high-resolution images. To address this, we replace the cross-attention blocks in the middle stage and the first block of the decoder with \textbf{Large Kernel Attention (LKA)}. These specific locations represent the transition where the model begins synthesizing global spatial coherence from compressed representations. LKA factorizes large-scale convolutions into spatial local, spatial long-range, and channel-wise element to expand the receptive field without prohibitive computational costs. The training objective is defined as:
\begin{equation}
\mathcal{L}_{\text{RestorePath}} = \mathbb{E}_{z, m, c, t, \epsilon} [ \| \epsilon - \epsilon_\theta(z_t, z_{\text{m}}, m, t, c) \|_2^2 ]
\end{equation}
where $z_t$ is the noisy latent at timestep $t$ and $\epsilon$ is the Gaussian noise.

\subsection{Inference}
\label{met:sec 2}
\noindent \textbf{Random Patch Diffusion with Distance Weighted Interpolation.} To restore artifacts reaching megapixel scales while maintaining global histological coherence, we employ a \textbf{Random Patch Diffusion} strategy. Let $J \in \mathbb{R}^{P_H \times P_W \times 3}$ represent a high-resolution image divided into a grid of uniform size $H \times H$. This division yields a grid of dimensions $N_H \times N_W$, where $N_H = P_H / H$ and $N_W = P_W / H$. We define $(k, l)$ as the discrete grid coordinates such that $0 \le k < N_H$ and $0 \le l < N_W$. A feature embedding $y_{k,l}$ is extracted from each grid cell using a PFM. A valid set of grid positions $\mathcal{S}$ is defined as $\{ (k, l) \mid r_{k,l} \le \tau \}$, where $r_{k,l}$ represents the artifact ratio in the corresponding grid cell and $\tau$ is the threshold to determine artifact-free regions. During the inference process, a patch of size $H \times H$ is randomly sampled at a continuous spatial coordinate $p = (y,x)$ within $J$. To guarantee the model obtains adequate morphological context despite the complete obstruction of the sampled patch by artifacts, we calculate a localized conditioning vector $c_p$ employing \textbf{Distance Weighted Interpolation (DWI)}. This vector is derived as a weighted summation of the valid embeddings in $\mathcal{S}$:
\begin{equation}
c_p = \frac{\sum_{(k,l) \in \mathcal{S}} w((k,l), p) \cdot y_{k,l}}{\sum_{(k,l) \in \mathcal{S}} w((k,l), p)}
\end{equation}
The spatial weight $w((k,l), p)$ reflects the proximity of the sampled patch to surrounding healthy tissue and is defined as:
\begin{equation}
w((k,l), p) = \frac{1}{\| (k \cdot H + \frac{H}{2}, l \cdot H + \frac{H}{2}) - p \|_2^\alpha + \delta}
\end{equation}
In this formulation, $(k \cdot H + \frac{H}{2}, l \cdot H + \frac{H}{2})$ represents the center coordinates of grid cell $(k,l)$, $\alpha$ is a distance attenuation factor, and $\delta$ is a small constant used to avoid division by zero. By propagating feature embeddings from valid neighboring regions into artifact-heavy zones, RestorePath ensures that the inpainted histological patterns remain contextually and biologically consistent with the global histological environment.

\noindent \textbf{Adaptive Guidance Scale.} Restoration quality is highly dependent on the guidance scale. We propose an \textbf{Adaptive Guidance Scale (AGS)} to dynamically modulate influence of the condition $c$ based on the patch content. We quantify image complexity by calculating the Laplacian Variance ($V_{\text{lap}}$) based on the artifact-free regions of the high-resolution image $J$ with respect to its artifact mask $M$. This complexity is then mapped to a base scale $s_{\text{base}} = f(V_{\text{lap}}(J, M))$ ranging from $s_{min}$ to $s_{max}$ to reflect intrinsic histological details. The applied scale $s_t$ for each patch is dynamically determined by its local artifact ratio $r_{\text{patch}}$. We define $k$ as the artifact occupancy threshold that distinguishes between the inpainting and generation modes.

\begin{equation}
s_t = \begin{cases} s_{\text{base}} & \text{if } r_{\text{patch}} > k \\ 0 & \text{otherwise} \end{cases}
\end{equation}

The final noise prediction follows the classifier-free guidance formulation:

\begin{equation}
\tilde{\epsilon}_\theta = \epsilon_\theta(z_t, z_{\text{m}}, m, t, \emptyset) + s_t \cdot (\epsilon_\theta(z_t, z_{\text{m}}, m, t, c) - \epsilon_\theta(z_t, z_{\text{m}}, m, t, \emptyset))
\end{equation}

This mechanism prioritizes original pixel preservation in partially masked patches while enforcing strong contextual priors in heavily damaged regions. 



\begin{table}[t]
    \centering
    \caption{Quantitative comparison of inpainting performance across different datasets.}
    \label{tab:Table 1}
    \setlength{\tabcolsep}{0.7pt} 
    \begin{tabular}{lccccccccc}
        \toprule
        \multirow{2}{*}{Methods} & \multicolumn{3}{c}{TCGA BRCA} & \multicolumn{3}{c}{Camelyon16} & \multicolumn{3}{c}{BACH} \\
        \cmidrule(r){2-4} \cmidrule(lr){5-7} \cmidrule(l){8-10}
        & \makecell{LPIPS$_M$\\($\downarrow$)} & \makecell{FID$_C $\\($\downarrow$)} & \makecell{Emb\\Sim($\uparrow$)} & \makecell{LPIPS$_M$\\($\downarrow$)} & \makecell{FID$_C $\\($\downarrow$)} & \makecell{Emb\\Sim($\uparrow$)} & \makecell{LPIPS$_M$\\($\downarrow$)} & \makecell{FID$_C $\\($\downarrow$)} & \makecell{Emb\\Sim($\uparrow$)} \\
        \midrule
        LaMa & 0.4492 & 117.54 & 0.7995 & \textbf{0.3610} & 83.28 & 0.8858 & \textbf{0.4199} & 135.61 & 0.8000 \\
        CoordFill & 0.6829 & 235.22 & 0.6723 & $-$ & $-$ & $-$ & $-$ & $-$ & $-$ \\
        SDM & 0.4555 & 39.69 & 0.9103 & 0.3780 & 54.24 & 0.9155 & \underline{0.4390} & 53.21 & 0.9114 \\
        \midrule
        RP (w/o LKA) & \underline{0.4474} & \underline{32.18} & \underline{0.9258} & \underline{0.3732} & 45.91 & 0.9289 & 0.4414 & 47.33 & 0.9135 \\
        RP (w/o DWI) & 0.5920 & 110.48 & 0.8225 & 0.3757 & \underline{44.72} & \underline{0.9304} & 0.4422 & \textbf{44.03} & \textbf{0.9232} \\
        \textbf{RestorePath} & \textbf{0.4467} & \textbf{29.66} & \textbf{0.9317} & 0.3754 & \textbf{44.57} & \textbf{0.9308} & 0.4418 & \underline{44.37} & \underline{0.9201} \\
        \bottomrule
    \end{tabular}
\end{table}

\section{Experiments}

\subsection{Datasets and implementation details}
We used three breast pathology datasets: TCGA BRCA, Camelyon16, and BACH. For all datasets, we extracted training patches at a resolution of $256 \times 256$ pixels ($20\times$ magnification) and utilized the UNI~\cite{UNI} PFM for feature embedding. Training masks were equally sampled from GrandQC~\cite{GrandQC} and \textit{failure modes}~\cite{PFM}, including artifacts such as markings, tissue folding, bubbles, and tissue tears.

\noindent \textbf{TCGA BRCA.} We utilized 1,105 WSIs for which GrandQC masks are available, partitioning them into 884 for training and 221 for testing. Training patches were extracted exclusively from artifact-free regions as defined by GrandQC. For validation, we selected 620 regions ($2048 \times 2048$ pixels) from the test set, with synthetic artifacts generated via \textit{failure modes} to match the training distribution.

\noindent \textbf{Camelyon16.} This dataset consists of 110 tumor and 160 normal cases. To address the Needle-in-a-haystack nature, tumor patches were extracted with a 32-pixel overlap within lesions, while normal regions were 20\% subsampled. The test set consists of 38 tumor regions from 20 small-lesion cases and 20 normal regions ($2048 \times 2048$pixels), all masked with synthetic bubbles to ensure uniform occlusion of the pathological context.

\noindent \textbf{BACH.} This dataset comprises 400 images across four classes: Normal, Benign, In Situ, and Invasive. We assigned 80 images per class for training and 20 for testing. Training patches were extracted with a 32-pixel overlap. Test artifacts were restricted to bubbles to evaluate the restoration of pathological context.

\noindent \textbf{Implementation Details.}
To ensure dataset-specific optimization, we trained separate models for each dataset using hyperparameters according to the LRDM~\cite{LRDM} framework. In addition, we set $H=256$, $\tau=0.3$, $\alpha=2$, and $\delta=10^{-6}$. For the AGS, $s_{min}$, $s_{max}$, and $k$ are assigned values of 2, 7, and 0.99, respectively.

\begin{table}[t]
    \centering
    \caption{Performance of MIL models on the Camelyon16 needle-in-a-haystack cases.}
    \label{tab:Table 2}
    \footnotesize 
    \setlength{\tabcolsep}{1.2pt} 
    \begin{tabular}{lcccccc}
        \toprule
        \multirow{2}{*}{Methods} & \multicolumn{2}{c}{ABMIL} & \multicolumn{2}{c}{TransMIL} & \multicolumn{2}{c}{CLAM$_{\text{MB}}$} \\
        \cmidrule(r){2-3} \cmidrule(lr){4-5} \cmidrule(l){6-7}
        & Acc. & AUC & Acc. & AUC & Acc. & AUC \\
        \midrule
        Origin & 85.0 $\pm$ 1.6 & 89.6 $\pm$ 2.6 & 81.5 $\pm$ 5.6 & 86.2 $\pm$ 2.9 & 84.5 $\pm$ 2.4 & 88.5 $\pm$ 3.1 \\
        \midrule
        Artifact & 73.5 $\pm$ 4.6 & 82.3 $\pm$ 2.7 & 67.5 $\pm$ 6.1 & 77.2 $\pm$ 4.6 & 72.5 $\pm$ 5.7 & 83.3 $\pm$ 3.4 \\
        D\&D & 55.0 $\pm$ 2.2 & 70.8 $\pm$ 5.0 & 53.5 $\pm$ 2.5 & 59.4 $\pm$ 6.2 & 53.0 $\pm$ 1.9 & 71.6 $\pm$ 4.2 \\
        \textbf{Restored} & \textbf{84.0 $\pm$ 1.2} & \textbf{90.3 $\pm$ 1.9} & \textbf{78.5 $\pm$ 6.0} & \textbf{85.0 $\pm$ 3.1} & \textbf{83.0 $\pm$ 3.3} & \textbf{90.3 $\pm$ 2.1} \\
        \bottomrule 
    \end{tabular}
\end{table}

\begin{figure}[t!]
\centering
\includegraphics[width=1\textwidth]{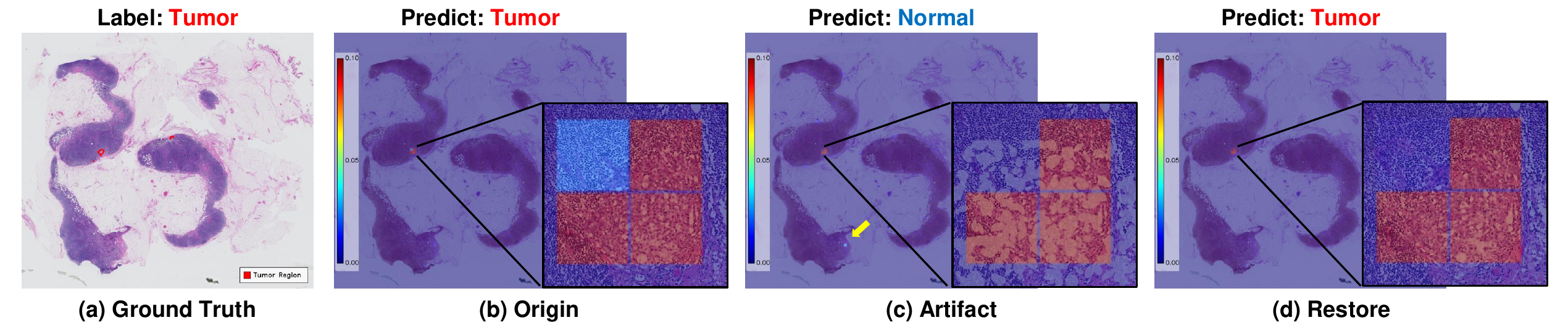}
\caption{MIL attention heatmaps on Camelyon16. (a) Ground Truth, (b) original, (c) artifact, and (d) restored images. Artifacts (c) cause attention dispersion (yellow arrow) and a silent failure (misclassified as Normal). RestorePath successfully refocuses attention on the lesion, ensuring accurate tumor prediction.}
\label{fig:Fig 4}
\end{figure} 

\subsection{Performance evaluation}

We evaluate quality via LPIPS$_{\text{M}}$, FID$_{\text{C}}$, and $\text{Emb Sim}$. LPIPS$_{\text{M}}$ is the LPIPS metric applied specifically to masked images. FID$_{\text{C}}$ replaces standard Inception-based FID with a CONCH PFM to capture pathological distributions more effectively. $\text{Emb Sim}$ measures the cosine similarity between CONCH features to verify diagnostic preservation.
RestorePath is compared against high-resolution image inpainting models including LaMa~\cite{LaMa}, CoordFill~\cite{CoordFill}, and an SDM~\cite{SDM} inpainting model applied with a random patch diffusion. Qualitative evaluations also demonstrate that RestorePath provides the most globally consistent results (Fig.~\ref{fig:Fig 2}). Ablation studies evaluating LKA and DWI are summarized in Table~\ref{tab:Table 1}. 

\noindent \textbf{Discussion.} Our framework achieves state-of-the-art performance in FID$_C$ and Emb Sim, confirming superior preservation of pathological features. While RestorePath leads in LPIPS for TCGA-BRCA, LaMa excels on Camelyon16 and BACH. This discrepancy stems from artifact shapes. TCGA-BRCA features opaque artifacts like markings and folds that entirely obscure underlying structures, making the context-aware interpolation of DWI essential. Conversely, air bubbles in other datasets allow localized textures to remain visible through gaps, favoring texture-based models like LaMa. Consequently, RestorePath is most impactful when surrounding context is the primary source of information.

\subsection{Downstream tasks}

\noindent \textbf{Multiple Instance Learning (Camelyon16).} We evaluated lymph node metastasis prediction using ABMIL~\cite{ABMIL}, TransMIL~\cite{TransMIL}, and CLAM~\cite{CLAM} with feature embeddings extracted from $256 \times 256$ pixel patches at $10\times$ magnification.  After 5-fold cross-validation on the training WSIs, models were tested on artifact-affected and restored WSIs. For D\&D strategy affected patches were zero-filled (black). For ABMIL, attention heatmaps were generated to visualize the restoration effect (Fig~\ref{fig:Fig 4}). While Artifact and D\&D cases showed significant drops from the Origin, RestorePath achieved performance within the standard deviation of the Origin slides, with ABMIL and CLAM even showing slight improvements in AUC (Table~\ref{tab:Table 2}). This validates restoration as superior to conventional strategies.

\begin{figure}[t!]
\centering
\includegraphics[width=1\textwidth]{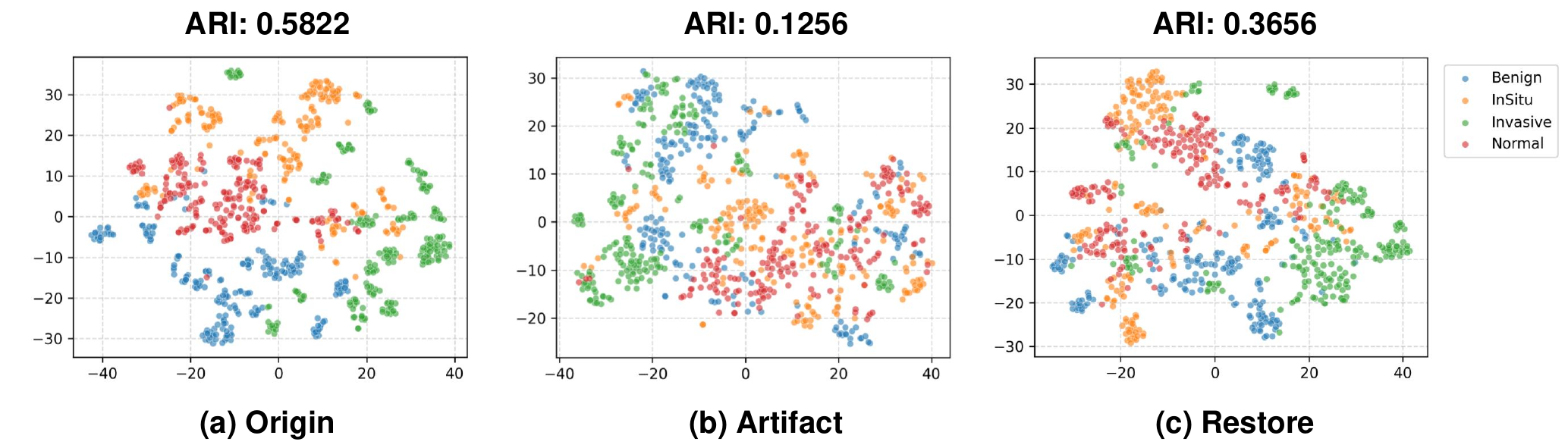}
\caption{t-SNE and ARI analysis on the BACH. (a) Origin: Baseline clusters with $ARI=0.5822$. (b) Artifact: Reduced class separability with $ARI=0.1256$. (c) Restore: RestorePath improves the $ARI$ to $0.3656$ by mitigating artifact-induced distortions.}
\label{fig:Fig 5}
\end{figure}

\noindent \textbf{Image Classification (BACH).} We trained ResNet50~\cite{ResNet}, DenseNet121~\cite{DenseNet}, and MobileNet-v2~\cite{MobileNetV2} via 5-fold cross-validation on $256 \times 256$ patches. Hard-voting inference was performed across Origin, Artifact, D\&D, and Restoration sets.  The t-SNE analysis revealed that restoration significantly recovered class separability. Adjusted Rand Index (ARI) scores improved from $0.1256$ in the artifact set to $0.3656$ after restoration, which is a significant recovery toward the origin score of $0.5822$ as illustrated in Fig~\ref{fig:Fig 5}. RestorePath recovers nearly half of the performance loss observed in Artifact and D\&D cases. This maintains superior diagnostic integrity compared to traditional QC procedures (Table~\ref{tab:Table 3}).

\begin{table}[t]
    \centering
    \caption{Performance of classification model on the BACH ($2048 \times 1536$) datasets.}
    \label{tab:Table 3}
    \footnotesize
    \setlength{\tabcolsep}{1.2pt}
    \begin{tabular}{lcccccc}
        \toprule
        \multirow{2}{*}{Methods} & \multicolumn{2}{c}{ResNet50} & \multicolumn{2}{c}{DenseNet121} & \multicolumn{2}{c}{MobileNetv2} \\
        \cmidrule(r){2-3} \cmidrule(lr){4-5} \cmidrule(l){6-7}
        & Acc. & AUC & Acc. & AUC & Acc. & AUC \\
        \midrule
        Origin & 86.2 $\pm$ 1.8 & 96.0 $\pm$ 0.3 & 86.8 $\pm$ 2.6 & 96.4 $\pm$ 0.6 & 87.0 $\pm$ 1.7 & 96.5 $\pm$ 0.1 \\
        \midrule
        Artifact & 61.0 $\pm$ 3.7 & 88.3 $\pm$ 1.1 & 68.0 $\pm$ 7.1 & 90.6 $\pm$ 2.6 & 73.2 $\pm$ 4.7 & 91.2 $\pm$ 0.6 \\
        \addlinespace[3pt]
        D\&D & 64.2 $\pm$ 4.1 & 89.3 $\pm$ 1.1 & 69.8 $\pm$ 7.8 & 91.5 $\pm$ 2.6 & 73.0 $\pm$ 3.9 & 92.1 $\pm$ 0.7 \\
        \addlinespace[3pt]
        \textbf{Restore} & \textbf{75.5 $\pm$ 1.3} & \textbf{94.4 $\pm$ 0.7} & \textbf{78.0 $\pm$ 2.8} & \textbf{94.0 $\pm$ 0.7} & \textbf{79.8 $\pm$ 2.2} & \textbf{93.9 $\pm$ 0.7} \\
        \bottomrule 
    \end{tabular}
\end{table}

\section{Conclusion}
We propose RestorePath, a diffusion-based patch-to-global framework for globally consistent megapixel-scale artifact restoration in WSIs. Our model based on inpainting LDM with LKA blocks conditioned on PFM embeddings. For high-resolution processing, we employ random patch diffusion with DWI and AGS for context-aware restoration leveraging surrounding artifact-free tissue. Benchmarks and downstream experiments demonstrate state-of-the-art performance, outperforming both the direct use of artifact images and conventional D\&D strategies. RestorePath serves as an advanced quality control tool to ensure diagnostic integrity in computational pathology.

%
\bibliographystyle{splncs04}
\bibliography{RestorePath}

\end{document}